# A combined Approach Based on Fuzzy Classification and Contextual Region Growing To Image Segmentation

Mahaman Sani Chaibou, Karim Kalti, Mohamed Ali Mahjoub
SAGE Research Unit ENISo, University of Sousse, 264 Erriadh, Sousse, TUNISIA
{sallaoudt@gmail.com, karim.kalti@gmail.com, medali.mahjoub@ipeim.rnu.tn}

**Abstract**

*We present in this paper an image segmentation approach that combines a fuzzy semantic region classification and a context based region-growing. Input image is first over-segmented. Then, prior domain knowledge is used to perform a fuzzy classification of these regions to provide a fuzzy semantic labeling. This allows the proposed approach to operate at high level instead of using low-level features and consequently to remedy to the problem of the semantic gap. Each over-segmented region is represented by a vector giving its corresponding membership degrees to the different thematic labels and the whole image is therefore represented by a Regions Partition Matrix. The segmentation is achieved on this matrix instead of the image pixels through two main phases: focusing and propagation. The focusing aims at selecting seeds regions from which information propagation will be performed. The propagation phase allows to spread toward others regions and using fuzzy contextual information the needed knowledge ensuring the semantic segmentation. An application of the proposed approach on mammograms shows promising results.*

*Keywords*--- **Image Segmentation, Fuzzy Classification, Region-growing, Context Information, Contextual Region-growing**.

## 1. Introduction

Image segmentation is a task of the image processing field that consists of partitioning an image into a set of homogenous regions. There are several image segmentation techniques proposed in the literature among which the region-growing technique [1-6]. Region Growing is an approach to image segmentation which is composed of two main steps: the seeds selection and the region expansion by merging with connected regions. One of the main drawbacks of approaches in this group is that both seeds selection and regions growing are based on regions intrinsic low-level features. Although, there are many interesting works in the region-growing topic, segmentation approaches in that group are still very different comparatively to the human vision system. The main reason is that the human vision uses high-level knowledge relative to the image [8]. In fact, by using only low-level features, these segmentation approaches can lead to merge regions that do not belong to the same object at semantic level.

We present in this work, a contextual region-growing approach to image segmentation combined with a semantic fuzzy classification of regions. The proposed approach uses prior contextual knowledge about regions in both seeds selection and regions growing steps. In fact, seeds selection is performed by regions fuzzy classification according to a set of thematic classes provided in the prior contextual knowledge. To allow semantic and coherent regions merging, we introduce a new region fuzzy context-based characterization, called Context Configuration. Merged regions features are updated from region intrinsic and context information.

The rest of this paper is organized as follows: the proposed segmentation approach is detailed in section 2, and its results are presented and discussed in section 3. Section 4 gives conclusions on the presented work.

## 2. Materials and methods

The main objective of this work is to propose an image segmentation approach that takes advantage of prior knowledge about regions. The proposed approach belongs to the general framework of region growing segmentation approaches. It presents two main contributions:

- The use of high-level features for segmentation: Common image segmentation approaches usually use low-level features. In this work, we use prior high-level information about regions to provide a semantic segmentation.
- A partly imitation of the human vision system behavior by focusing and knowledge propagation. Indeed, to analyze image, humans generally identify regions that are easily recognizable. Then they use knowledge about those regions to recognize the rest of the image. In our approach, we reproduce those steps by a focusing step dedicated to identify seeds and a propagation step that allow spreading contextual information from seeds to their connected neighbor regions.

The proposed approach presents two main phases: an initialization phase and an iterative phase that combines region fuzzy classification and region-growing.

## 2.1. Initialization

This phase prepares the input image for the region-growing phase. It is divided into three main steps: a knowledge acquisition, an over-segmentation of the image and finally a classification of the regions provided by the over-segmentation based on thematic regions classes.

The knowledge acquisition step is devoted to gather prior contextual knowledge about regions. In image segmentation, the contextual knowledge generally refers to thematic regions classes and their spatial relationships. From this information, we define the regions valid configurations as possible regions spatial arrangement allowed in the image domain. Those configurations will later be used to assess region classification and information update. The provided thematic regions classes set, let say $C = \{C_1, C_2, ..., C_K\}$, will be used for classification of image regions at the segmentation phase.

The over-segmentation step produces the initial set of regions $R = \{R_1, R_2, ..., R_N\}$ from which the region-growing will begin. Any over-segmentation algorithm can be used for this step, such as watershed [9, 10], or super-pixels [11, 12].

We perform a classification of the regions obtained from the over-segmentation step. This allows injecting prior information about regions in the segmentation process. This step will give a fuzzy membership degrees vector (MDV) $M_k$ for each class $C_k$. From this point, all operations will be based on membership degrees.

## 2.2. Iterative combination of contextual classification and region-growing

In this phase, we model the process of focusing, knowledge propagation partly imitating the human vision behavior to image analysis. To this end, we propose the four following steps:

- Focusing: to identify seed regions
- Contextual configuration identification: to capture spatial relationship of regions and their context
- Knowledge propagation: to spread information from seeds to non-seed regions
- Conditional merging: to eventually merge connected similar regions.

The phase begins with the selection of seed regions by the focusing step. In order to propagate information from seeds, spatial relationships of regions and their context are identified from those already defined in the initialization phase. According to regions configurations, we defined a fuzzy update scheme to actualize non-seed regions information based on their context to propagate seeds information. Updated regions are reclassified to let them propagate their acquired information to their neighbors. Those three steps are iterated until a stability of regions information is reached.

After stability, we propose to merge similar direct neighbor regions and the approach restarts from the focusing step until convergence. Delaying the merger of regions as much as possible, we reduce the false classification of regions.

### 2.2.1. Focusing

Seeds selection is a crucial step in region-growing segmentation approaches. Indeed the result of the segmentation is highly dependent on the initial seeds from which regions are expanded. A second fuzzy classification of $R$ into three classes according to the results of the first classification is performed to provide support for seeds selection and region merging steps. For each region $R_i$, we define the following terms:

The membership degrees:
$$MD_i = \{MD_{i,1}, MD_{i,2}, ..., MD_{i,K}\} \quad (1)$$
as the set of its membership degree for the different classes $C_k$

The sorted membership degrees set:
$$SMD_i = \{SMD_{i,1}, SMD_{i,2} ..., SMD_{i,k}\} \quad (2)$$
The maximum membership degree:
$$MMD_i = \max(MD_i) \quad (3)$$
The predominant class PC of a region $R_i$:
$$PC(R_i) = C_k; \ MMD_i = MD_k \quad (4)$$
Membership distribution gap:
$$DG_i = \{g_{i,1}, g_{i,2}, ..., g_{i,k-1}\} \quad (5)$$
where $g_{i,j} = MD_{i,j+1} - SMD_{i,j}$

To reduce the approach dependence on parameters, and avoid threshold, we classify the $DG_i$ into three equal partitions and define the Separation Coefficient of a $MD_{i,j}$ as:

$$SC(MD_{i,j}) = \frac{1}{\max(DG_i)} \quad (6)$$

An unsupervised fuzzy classification of R into three classes according to their predominant class (PC) is performed. Thus, we have regions with high classification degree (HCD), regions with Medium Classification Degree (MCD) and regions with Low Classification Degree (LCD).

$$class(R_i) = \begin{cases} HCD & if \ SC(MMD_i) > 0.6 \\ MCD & if \ SC(MMD_i) \in [0.3, 0.6] \\ LCD & otherwise \end{cases} \quad (7)$$

In this approach, the seeds are the HCD regions.

After seeds selection, contextual information such as MD of the context regions and their spatial relationship are used to merge connected regions and update their membership degrees.

### 2.2.2. Contextual configuration identification

HCD regions do not change membership degrees during the update process. MCD regions' membership degrees are not final, they can change during updates. LCD regions cannot spread their local information to their neighbors but their membership degrees can be updated accordingly to their HCD or MCD direct neighbors.

Considering that the current region is $R_i$, its local context, denoted by LC, is composed of L regions neighbor to $R_i$. We define four main contextual configurations and their corresponding update rules, presented in table 1:

**Table 1: Contextual configuration base**

| Configuration | Update Rules |
|---|---|
| LC is homogeneous and similar to $R_i$ in low-level features. | Increase the degree of $R_i$ that represents the thematic class of LC. |
| LC is homogeneous and non-similar in low-level features to $R_i$ and all LC regions are HCD. | Increase the degree (in $MD_i$) of classes that may be included in the class LC. |
| LC is homogeneous and non-similar in low-level features to $R_i$ and some LC regions are MCD or LCD. | Increase the degree (in $MD_i$) of classes that may be neighbor with the class LC. |
| Heterogeneous LC. | Increase the degree of classes that form a valid configuration with LC classes. |

### 2.2.3. Knowledge propagation

The purpose of this step is to update the membership degrees of regions taking into account their neighborhood. For each of the previously defined configurations, the update is performed as a function of the difference $d_{R_i}$ between the region and its neighbors. The latter distance is equal to the weighted average of the distances between the region and each of its neighbors taken separately.

$$d_{R_i} = \frac{1}{L}\sum_{j=1}^{L} d(R_i, V_j) \quad (8)$$

The difference between two regions is computed based on the distribution of their respective membership degree. We use for that the Bhattacharyya distribution distance [13].

Updating degrees of a region consists in partitioning the membership degrees of that region into two parts: $m_1$ (for classes giving a valid configuration with $R_i$) and $m_2$ (with no valid configuration with $R_i$) and then increase membership degrees of $m_1$ by a value of $\varepsilon$:

$$\varepsilon = \frac{\frac{\sum_i a_i}{n}(1-d_R)}{K} \quad (9)$$

and reduce those of $m_2$ by $\frac{\varepsilon}{K-1}$

Where: $a_i \in m_1$ and $n = card(m_1)$

These values are set depending on the distance between the region and the local context.

Equation 10 gives the new membership degree values of $R_i$ after the update, when the membership degree of $R_i$ to $c_j$ is increased:

$$\begin{pmatrix} a_1 \\ \vdots \\ a_j \\ \vdots \\ a_K \end{pmatrix} = \begin{pmatrix} a_1 - \frac{a_j(1-d_R)}{K(K-1)} \\ \vdots \\ a_j + \frac{a_j(1-d_R)}{K} \\ \vdots \\ a_K - \frac{a_j(1-d_R)}{K(K-1)} \end{pmatrix} \quad (10)$$

### 2.2.4. Conditional region merging

The iterations of previous steps accumulate information in regions, which allow them to change their characteristics progressively and to tend to homogeneity with their neighborhood. After updating their information, some neighboring regions reach a similarity allowing them to merge. Membership degrees of the resulted regions are recomputed as the average of the merged regions. Thus, the merging phase, in which two regions are merged into one, is not systematic but only intervenes when its conditions are met. The latter conditions are defined through heuristics on information about regions such as membership degree-based similarity. Figure 1 gives the anagram of the segmentation phase.

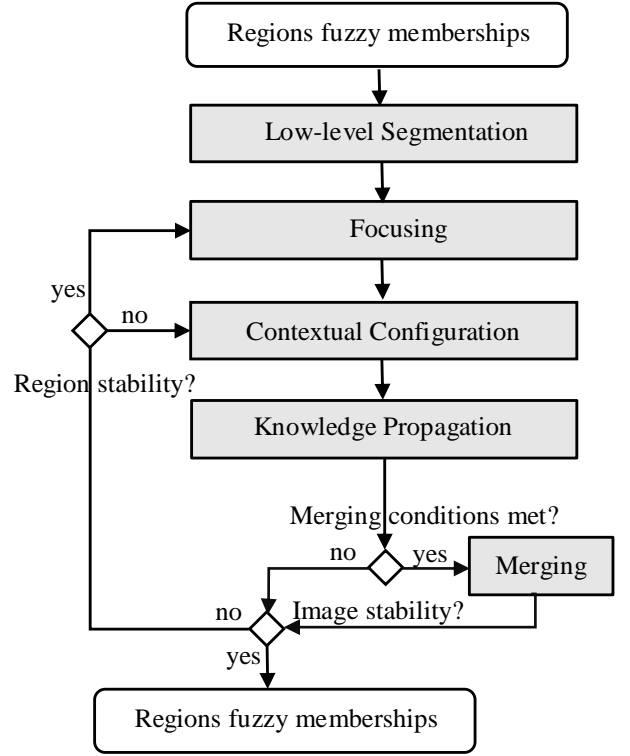

**Figure 1: Anagram of the processing phase**

To summarize, below we present the algorithm of the processing phase.

```
Algorithm : Segmentation
Inputs : InputImage, IBase, UpdateRules
BEGIN
1. Regions=overSegmentation (InputImage)
2. While( convergence(Regions)==false) do
3.   AppMatrix=classification(Regions)
4.   While(convergence(Regions)==false) do
5.     SeedRegions=focusing(Regions, AppMatrix)
       [Regions,AppMatrix]=InfUpdate(Regions,
          AppMatrix, IBase, UpdateRules)
       If(MergeConditions==true)
6.       [Regions, AppMatrix]=merge(Regions,
            AppMatrix)
7.     EndIf
8.   EndWhile
9. EndWhile
END
```

## 3. Result and discussion

### 3.1 Initialization

Image segmentation by human strongly depends on what the knowledge he/she had on the image. Our approach begins by collecting information about the considered image to process. For this purpose, we use regions and their spatial relationship (neighborhood and inclusion) as domain information. Considering a mammogram, like in figure 2, the collected information could be as in table 2:

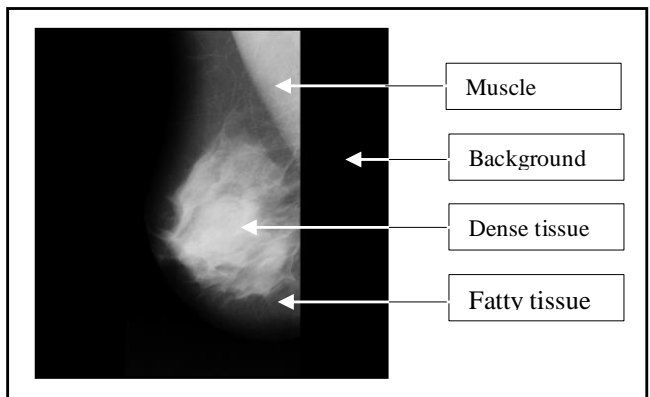

**Figure 2: example of information on mammogram image**

**Table 2: Domain contextual information base**

| Regions | Spatial Relationship | |
|---|---|---|
| | *Neighbor* | *Included* |
| Background | Muscle, Fatty tissue | --- |
| Muscle | Background, fatty tissue | --- |
| Fatty tissue | Background, Muscle, Opacity | Dense tissue |
| Dense tissue | Muscle, Fatty tissue | --- |

From the domain information base, the approach build the set of possible configuration for regions in a mammogram base. A configuration stands for a given region and a valid set of neighboring regions (local context). Table 3 gives all valid regions spatial distribution from the previous domain information.

**Table 3: Contextual configurations**

| Regions | Local Context |
|---|---|
| Background | Muscle, Fatty tissue |
| Muscle | Background, Fatty tissue |
| Fatty tissue | Background, Muscle |
| Fatty tissue | Background, Dense tissue |
| Fatty tissue | Muscle, Dense tissue |
| Fatty tissue | Background, Muscle, Dense tissue |
| Dense tissue | Fatty tissue |
| Dense tissue | Muscle, Fatty tissue |

### 3.2 Region-growing

In this phase, the gathered information will be used to segment mammograms. First, an over-segmentation algorithm provides an initialization set of regions from which computations will begin. Figure 3 gives the input image and the over-segmentation result.

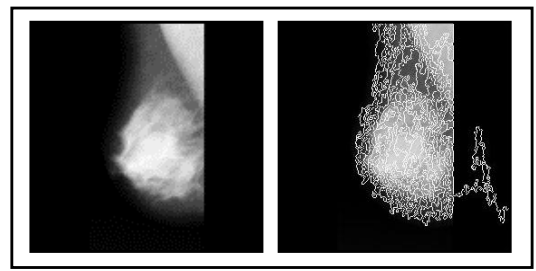

**Figure 3: input image and initialization regions set**

A fuzzy classification of the initialization regions set into four classes (namely Background, Muscle, Breast fatty tissue and Breast dense tissue) provides four membership degrees matrices, one for each class. Each membership degree *MD* contains the probability for each region to belong to the corresponding class. The sum of probabilities of a region of belonging to each class is equal to one. Details about the evolution of the fuzzy classification and the region information update on the image are shown in figure 4.

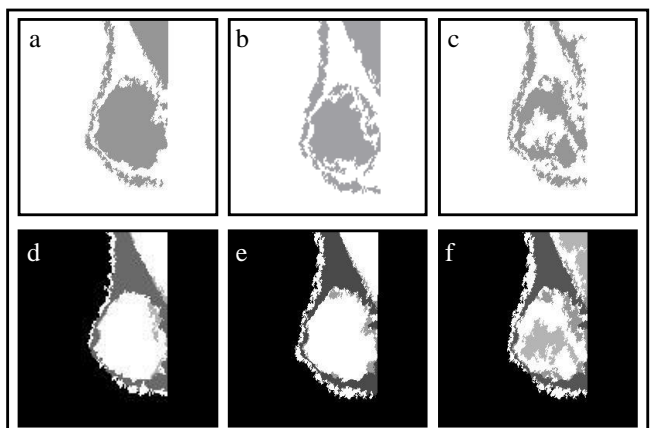

4-a, 4-b and 4-c present regions classification quality respectively before the region-growing process starts, at 100$^{th}$ loop and at 200$^{th}$ loop of the region-growing. In those images, white color are HCD regions, gray are MCD regions and black are LCD regions.
4-d: Labeled regions before region-growing process; 4-e: Labeled regions at 100$^{th}$ loop of region-growing; 4-f: Labeled regions at 200$^{th}$ loop of region-growing.

**Figure 4: A preview of the region information update process.**

At the end of the process a defuzzification step is used to retrieve the segmented image. In this step, we present two choices:

- Only HCD regions are accepted and so replaced by the corresponding regions classes, whereas LCD and MCD regions are represented by white areas i.e. are unlabeled.

- HCD and MCD regions are accepted and only LCD regions are still unlabeled in the segmented image.

The input image and its segmentations (for HCD only and for HCD & MCD) are presented in the figure 5. Unlabeled regions are shown in white color on the segmented images.

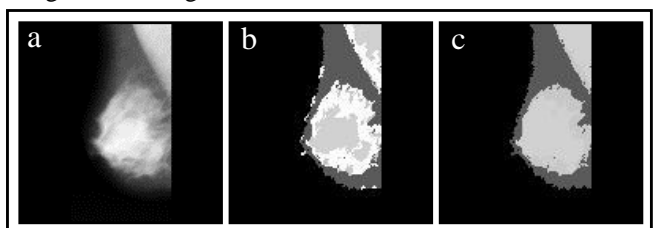

5-a:original image; 5-b: segmented with only HCD regions accepted; 5-c: segmented image with HCD and MCD regions accepted

**Figure 5: input image and segmentation result.**

In the figure 4, images a to c, show the evolution of regions classification which is used to produce the segmented image. At the beginning (image a) almost all muscle and breast dense tissue are classified as MCD regions but as the process iterates some regions of those parts of the image become HCD. As shown by the segmented images (fig. 5, images b & c) when de considering only HCD regions in defuzzification, image segmentation does not cover all regions. Adding MCD regions in defuzzification improves the segmentation but this induces the use of regions with an average quality of classification.

## 4. Conclusions

In this work, we presented an image segmentation approach that combines between a semantic fuzzy classification of regions and a context-based region-growing. The semantic classification allows to efficiently characterizing region based on prior domain information. An over-segmentation of the image is first performed and obtained regions are labeled according to the prior information. Based on the results of this first classification, a second classification of regions into LCD, MCD and HCD regions allows to select HCD regions as the seeds for the region-growing phase. Hence, LCD and MCD regions with HCD regions in their context will update their information according to their context. The update allow to changing membership degrees of current regions, such that they hopefully become HCD.

The proposed approach provides a contextual fuzzy classification that allows to revise regions assignment by updating their information. This is a key point that ensures a domain specific information homogenous image segmentation. Moreover, the approach is adaptive to various application domain since it allows dynamic context information integration.